\documentclass[journal]{IEEEtran}
\ifCLASSINFOpdf
\else
\fi
\usepackage{amssymb}
\usepackage{graphicx}
\usepackage{amsmath}
\usepackage{caption,setspace}

\newcommand{\eqnref}[1]{Eq.~(\ref{#1})}

\begin{document}
%
\title{Expert Level control of Ramp Metering based on Multi-task \\ Deep Reinforcement Learning}


\author{\IEEEauthorblockN{Francois Belletti\IEEEauthorrefmark{1},
Daniel Haziza\IEEEauthorrefmark{2},
Gabriel Gomes\IEEEauthorrefmark{3},
Alexandre M. Bayen\IEEEauthorrefmark{4},~\IEEEmembership{Member,~IEEE}}
\IEEEauthorblockA{\IEEEauthorrefmark{1}Computer Science Dept.,
UC Berkeley, 94720, USA}
\IEEEauthorblockA{\IEEEauthorrefmark{2}Ecole Polytechnique, Palaiseau, 91128, France}
\IEEEauthorblockA{\IEEEauthorrefmark{3}
Inst. of Transportation Studies,
UC Berkeley, 94720, USA}
\IEEEauthorblockA{\IEEEauthorrefmark{4}EECS Dept., CEE Dept. and
Inst. of Transportation Studies,
UC Berkeley, 94720, USA}
}

%



\IEEEtitleabstractindextext{%

\begin{abstract}
    This article shows how the recent breakthroughs in Reinforcement Learning (RL) that have enabled robots to learn to play arcade video games, walk or assemble colored bricks, can be used to perform other tasks that are currently at the core of engineering cyberphysical systems. We present the first use of RL for the control of systems modeled by discretized non-linear Partial Differential Equations (PDEs) and devise a novel algorithm to use non-parametric control techniques for large multi-agent systems.
    Cyberphysical systems (e.g., hydraulic channels, transportation systems, the energy grid, electromagnetic systems) are commonly modeled by PDEs which historically have been a reliable way to enable engineering applications in these domains. However, it is known that the control of these PDE models is notoriously difficult. 
    We show how neural network based RL enables the control of discretized PDEs whose parameters are unknown, random, and time-varying. We introduce an algorithm of Mutual Weight Regularization (MWR) which alleviates the curse of dimensionality of multi-agent control schemes by sharing experience between agents while giving each agent the opportunity to specialize its action policy so as to tailor it to the local parameters of the part of the system it is located in.
    A discretized PDE such as the scalar Lighthill–-Whitham-–Richards (LWR) PDE can indeed be considered as a macroscopic freeway traffic simulator and which presents the most salient challenges for learning to control large cyberphysical system with multiple agents.
    We consider two different discretization procedures and show the opportunities offered by applying deep reinforcement for continuous control on both.
    Using a neural RL PDE controller on a traffic flow simulation based on a Godunov discretization of the of the San Francisco Bay Bridge we are able to achieve precise adaptive metering without model calibration thereby beating the state of the art in traffic metering. Furthermore, with the more accurate BeATS simulator we manage to achieve a control performance on par with ALINEA, a state of the art parametric control scheme, and show how using MWR improves the learning procedure.
\end{abstract}

}

\maketitle

\IEEEdisplaynontitleabstractindextext

%
\IEEEpeerreviewmaketitle

\section{Introduction}
%
%
%
%
\IEEEPARstart{I}{n} the United States alone, the cost of direct and indirect consequences of traffic congestion was estimated to 124 billions USD in 2013, this cost taking the form of time spent by commuters in traffic jam, air pollution, accidents, etc. It represents almost 1\% of the country's GDP, and is expected to grow by 50\% within the next 15 years. Dealing with this issue is becoming a priority of government agencies, as the U.S. Department of Transportation Budget rose to almost 100 billions USD in 2016. In this context, any improvement to travel times on highways can lead to tremendous nationwide and worldwide improvements for the economy and the environment.

Maintaining and building road infrastructure, as well as urban planning are the most obvious ways to adapt the traffic network to the ever growing demand for mobility. However, changing the network architecture can only occur seldom and at an expensive cost.

Control of traffic flow is an alternate approach to addressing this issue as it aims at using existing infrastructure more efficiently and adapt it dynamically to the demand. In this article, we introduce new techniques for traffic control based on advances in Reinforcement Learning (RL) and Neural Networks. As opposed to most commonly used approaches in traffic control, we want to achieve control in a model-free fashion, meaning that we do not assume any prior knowledge of a model or the parameters of its dynamics, and thus do not need to rely on the expensive and time consuming model calibration procedures.

Recent developments in Reinforcement Learning (RL) have enabled machines to learn how to play video games with no other information than the screen display \cite{mnih2013playing}, remarkably beat champion human players at Go with the AlphaGo program \cite{alphago}, or complete various tasks including locomotion and simple problem solving \cite{levine2015end,finn2015deep,duan2016benchmarking}. The advent of policy gradient-based optimization algorithms enabling RL for highly-dimensional continuous control systems as can be found in \cite{levine2013guided,schulman2015high}, has generalized model-free control to systems that were characteristically challenging for Q-learning. Q-learning approaches, although successful in \cite{mnih2013playing} suffer from a curse of dimensionality in continuous control if they rely on discretizing the action space.

In this article, RL trains a traffic management policy able to control the metering of highway on-ramps. The current state of the art Ramp-Metering policy ALINEA \cite{alinea} controls incoming flow in order to reach an optimal density locally. This optimal density depends on the model used and has to be manually specified to have an optimal control. Recently, non-parametric approaches based on Reinforcement Learning such as \cite{fares2014freeway} or \cite{rezaee2014decentralized} have been proposed to achieve ramp-metering, but face two main limitations. These methods are not scalable beyond a few on-ramps, and limit traffic management to on-ramp control.

We introduce a way to learn an optimal control policy with numerous agents, and demonstrate the flexibility of our approach by applying it to different scenarios.
The following contributions of our work are presented in this article:
\begin{itemize}
    \item We introduce a 
    framework to use RL as a generic optimization scheme that enables the control of discretized Partial Differential Equations (PDEs) in a robust and non-parametric manner. To our knowledge, this is the first use of RL for control of PDEs discretized by finite differencing. Discretized non-linear PDEs are notoriously difficult to control if the difference scheme used is non-smooth and non-continuous, which is usually required to capture nonlinear and non-smooth features of the solution (as is the case here).
    \item In the case of PDEs used to model traffic, we demonstrate on different examples an extensive control over boundary conditions as well as inner domain for the first time in a non-parametric way. We showcase the robustness of the approach and its ability to learn from real-world examples by assessing its performance after an extremely noisy training phase with stochastic variations in the underlying PDE parameters of the model used
    \item We introduce an algorithm to train Neural Networks that we denote Mutual Weights Regularization (MWR) which enables the sharing of experience between agents and specialization of each agent thanks to Multi Task Learning \cite{caruana1998multitask}. MWR is a Neural Network training approach that allows Reinforcement Learning to train a policy in a multi-agent environment without being hampered by a curse of dimensionality with respect to the number of agents. Applied to the actual traffic control problem of ``Ramp metering'', our model-free approach achieves a control of a comparable level to the currently used and model-dependent implementation of ALINEA which constitutes the state art and was in our case calibrated by an world-wide renowned traffic engineer.
\end{itemize}

We first present the PDEs used to simulate traffic and introduce the generic PDE discretization scheme. A first simulator based on a Godunov scheme \cite{Godunov} is used to demonstrate the efficiency of our approach on multiples situations. The Berkeley Advanced Transportation Simulator (BeATS), a state of the art macroscopic simulator implementing a particular instantiation of the Godunov scheme sometimes referred to as the Cell Transmission Model \cite{muralidharan2009freeway} and used in traffic research \cite{wan2013prediction} is also introduce, as we use it for our final benchmark. Traffic control is presented in the form of a Reinforcement Learning problem and we present the MWR algorithm to mitigate the issues arising from the high dimensionality of the problem. We eventually present the results we achieve, and compare our algorithm to the pre-existing state of the art techniques. This state of practice reference, an ALINEA control scheme calibrated by traffic engineers at California PATH can be considered representative of state of the art expert performance. To the best of our knowledge, we are the first to present a non-parametric scalable method calibrated with RL that performs as well as the pre-existing parametric solutions provided by traffic engineering experts.

\section{Models and methodology}

\subsection{Highway traffic PDE}
A highway vehicle density may be modeled by a Partial Differential Equation (PDE) of the following form:
\begin{equation}
    \tag{LWR PDE}
    \partial_t \rho + \partial_x F \left( \rho \right) = 0.
    \label{eq:LWR}
\end{equation}

For a given uniform vehicle density $\rho$, $F(\rho)$ is the flow of the vehicles (in vehicles per time unit); $F$ is the fundamental diagram, its maximum is called the critical density, and corresponds to the optimal density to maximize the flow.

$F$ usually has the following typical shape:
\begin{itemize}
    \item When the density $\rho$ is lower than the critical density $\rho_{crit}$, the vehicle flow increases with the density
    \item When the density exceeds this value, congestion appears, and adding more vehicles actually reduces the flow
\end{itemize}

\begin{center}
    \includegraphics[trim = 0mm 0mm 0mm 0mm, clip, width = 0.5\textwidth]{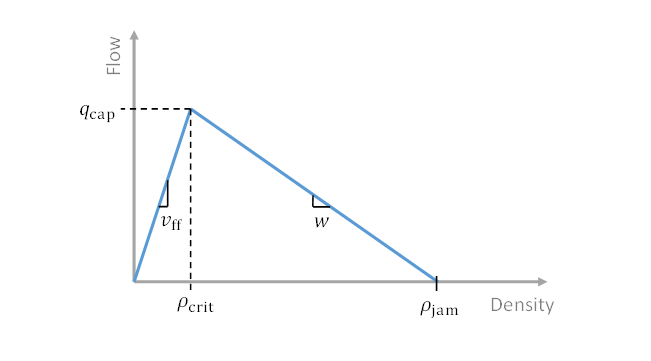}
	\captionof{figure}{Example fundamental diagram}
\end{center}

\subsection{Control of PDEs with reinforcement learning}
Non-parametric control of PDEs takes the form in the present article of a Markov Decision Process (MDP) which is formally introduced below in \ref{RL}.
The PDE in its discretized form devises a transition probability $P$ between two different states of the system.
Solving a MDP is a known procedure if the transition probability $P$ is known beforehand, using techniques such as Dynamic or Linear Programming. If $P$ is unknown, it is more challenging. However, it is more appealing a procedure as devising $P$ for discretized PDEs is generally intractable and requires estimates of the parameters of the system. Also, as we operate in a continuous action space we will not consider Q-learning based approaches which are typically challenged by high-dimensionality in that setting.
Therefore, policy gradient algorithms present a compelling opportunity  as model dynamics are sampled from the simulation trials in algorithms such as \cite{kakade2001natural, peters2006policy, van2015learning, zzzheess2015learning} and no prior knowledge of the model is necessary to train the policy. This creates a model-independent paradigm which abstracts out the model and makes the approach generic.\\

\subsection{Simulation}
The experimental method we followed consists of two steps.

A first one uses a coarser less realistic discretization scheme for the \ref{eq:LWR} named after its inventor: Godunov \cite{Godunov}. We provide more details about this scheme in Appendix \ref{sec:godunovDicr}. This step serves the purpose of a proof of concept that discretized PDEs can be controlled by a neural net trained using a RL scheme in completely different situations and with different objectives. Our first contribution is to show that policy gradient algorithms achieve that aim.

In our second step we use a more accurate cell transition model \cite{muralidharan2009freeway} scheme entailed in the state of the art BeATs simulator in order to show that the procedure we present is still valid in a more realistic setting. We also change the tasks we assigned to the control scheme in order to precisely account for the actual needs of traffic management systems used in production. We show that training our neural net based policy by policy gradient methods achieves comparable performance with the state of the art ALINEA control scheme \cite{alinea} although the former is non-parametric when the latter requires a calibration of traffic related parameters. In both cases, the neural net manages to implicitly learn the intrinsic properties of the road segment under consideration and provide a good control policy.


\section{Controlling cyberphysical systems with Neural Networks trained by Reinforcement Learning}
In order to be able to control a complex cyberphysical system a non-parametric manner, we adopt a Reinforcement Learning formulation.

\subsection{Reinforcement Learning formulation\label{RL}}
RL is concerned with solving a finite-horizon discounted Markov Decision Process (MDP). A MDP is defined by a tuple $\left(S, A, P, R, P_0, \gamma, T \right)$. The set of states is denoted $S$ and will typically be $\mathbb{R}^d$ in our instance where $d$ is the number of finite differences cells as in \cite{Godunov}. The action space $A$ will correspond to a vector in $\mathbb{R}^d$ which represents the vehicular flow that the actuator lets enter the freeway which corresponds in the present case to the weak boundary condition implemented in the form of a ghost cell. The transition probability $P: S \times A \times S \rightarrow \mathbb{R}_{+}$ 
is determined by the freeway traffic simulator we use i.e. the Godunov discretization scheme  and the stochastic queue arrival model devised, discussed below. Random events such as perturbations to the input flow of vehicles or accidents affect the otherwise deterministic dynamics of the discretized system. The transition probability $P$ is affected by these events and there likelihood but does not need to be known analytically for the system to operate nor be estimated. This is one of the key advantages offered by Reinforcement Learning over other approaches.
The real valued reward function $R$ is for the practitioner to define which implies that the same training algorithm can be used to achieve different objectives. The initial state distribution is denoted $P_0$, the discount factor $\gamma$ and the horizon $T$.
%
Generically, RL consists in finding the policy $\pi_\theta : S \times A \rightarrow \mathbb{R}_{+}$ that maximizes the expected discounted reward 

$$
\mathbb{E}\left(\eta(\pi)\right)
\:\text{ where }\:
\eta(\pi) := \sum_{t=0}^{T_{max}} \gamma^t r_t
$$
We denote $\tau = \left(s_0, a_0, \ldots\right)$ the representation of a trajectory defined by the probability distributions $s_0 \sim P_0$, $a_t \sim \pi(a_t | s_t)$, the state transition probability $s_{t+1} \sim P\left(s_{t+1} | s_t, a_t \right)$ and the reward distributioon $r_t \sim P\left(r_t|s_t,a_t\right)$.
We will consider a stochastic policy which defines a probability distribution of $a_t$ conditional to $s_t$ (or the observation of the state at time $t$) parametrized by $\theta$. This creates a stochastic regularization of the objective to maximize and enables to computation of the gradient of a policy with respect to its parameters in spite of the dynamics of the system under consideration not being differentiable, continuous, or even known.\\

\subsubsection{Reinforcement Learning based control of discretized PDEs}
The recent developments in RL featured in \cite{levine2014learning,schulman2015high,schulman2015trust} guarantee convergence properties similar to those of standard control methods and therefore strongly motivates their usage for the control of PDEs. On the contrary, since they are being model-independent, they are intrinsically robust to varying parameters and are able to track parameter slippage. This leads us to consider them as the new generation of generic control schemes.
We show how the use of RL on discretized PDEs enables the extension of schemes to systems featuring random dynamics, unknown parameters and regime changes, hence surpassing parametric control schemes.
%
The MPC approach in \cite{zzzbellemans2006model} and the adjoint method based technique of \cite{reilly2015adjoint} both rely on the definition of a cost function which needs to be minimized. For PDEs such as the LWR PDE, one typically maximizes throughput, minimizes delay, or a functional of the state (for example encompassing energy emissions). A RL approach will therefore focus on maximizing a decreasing function of that cost which will be our accumulated reward. This is standard practice to encode an operational objective.\\


\subsubsection{State and action space.}
Consider a discretized approximation of the solution $y$ to \eqnref{eq:conservationeq} (see appendix) by the Godunov scheme described in appendix. The solution domain is $\left[0, \: T\right] \times \left[0, \: L \right]$, the discretization resolutions $\Delta t$ and $\Delta x$ satisfying $\Delta t \leq c \Delta x$ are chosen to meet well posedness requirements (Courant-–Friedrichs–-Lewy condition where $c$ is the maximum characteristic speed of \eqnref{eq:conservationeq}). The solution to the equation is approximated by a piecewise constant solution computed at the discrete time-space points $\left\{0, \Delta t, \ldots, T - \Delta t , T\right\} \times \left\{0, \Delta x, \ldots, L - \Delta x , L\right\}$. The action space for this system  consists of incoming flow at the discretized elements $\left\{0, \Delta t, \ldots, T - \Delta t , T\right\} \times \left\{ 0 \right\}$, and generally belongs to a bounded domain $\left[0 \ldots u^{\text{max}_{i,j}}\right]$. The policy will control this vector of incoming flows at each time step.\\

\subsection{Neural Networks for Parametrized stochastic policies.}
In this paragraph we show how to construct an actuator based on a Neural Network.

\subsubsection{Parametrized stochastic policies.}
A vast family of stochastic policies are available for us to use so as to choose an action conditionally to an observation of the state. A common paradigm is to create a regression operator, typically a Neural Network, which is going to determine the values of the parameters of a probability distribution over the actions based on the space observation.
We practically use a Multilayer Perceptron that determines the mean and covariance of a Gaussian distribution over the action space. The action the policy undertakes is sampled from this parametrized distribution and will manage to maximize its expected rewards provided a reliable training algorithm is used.\\

\subsection{Neural Networks}
The policy we train are implemented as Artificial Neural Networks, containing Artificial Neural wired together.
\subsubsection{Artificial Neural Model}
For $p \in \mathbb{N}$, an Artificial Neural computes an output $y \in \mathbb{R}$ from an input vector $X \in \mathbb{R}^p$ with the following formula:
$$
y = \phi(W X + b)
$$
Where $\phi: \mathbb{R} \to \mathbb{R}$ is called the activation function which responds to the outcome of an affine transformation of its input space parametrized by $W \in \mathbb{R}^p$ and $b \in \mathbb{R}$.

\subsubsection{Networks}
A group of $q$ artificial neurons can be linked to a single input $X \in \mathbb{R}^p$ to create an output vector $Y \in \mathbb{R}^q$. This forms a neural layer.
When several layers are stacked, with the output of one being the input of the next one, one can speak of an Artificial Neural Network, whose input is the input of the first layer, and output is the output of the last layer.

The general organization and architecture of such a network may vary depending on usages, the type of input data to process. 
In the setting of computer vision, convolutional neural networks famously achieved human level image classification thanks to the invariance by translation and rotation of the convolution masks they progressively learnt \cite{krizhevsky2012imagenet}.

\begin{figure}[!t]
\centering
\includegraphics[trim={0 1.8in 0 1.2in},clip,width=3.5in]{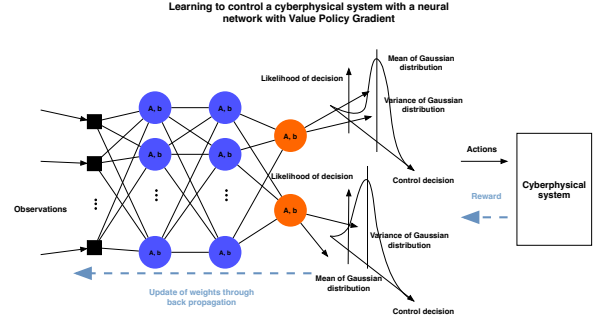}
\caption{Non parametric control learnt by experience. A neural network decides the parameter of probability distributions the actions will be sampled from based on observations of the state.}
\label{fig_sim}
\end{figure}

\textbf{Back propagation}
In order to train Neural Networks back-propagation \cite{ le1990handwritten} is a key algorithm that performs a stochastic gradient descent on a non-convex function \cite{bishop1995neural}.
Approaches to train such a Neural Network for control in the Q-learning framework has been adopted in \cite{mnih2013playing} and were successful in a discrete control setting. With continuous control, a different family of methods is generally used that encourage the policy entailed in the network parameters to take actions that are on average advantageous and discourage actions that have an expected negative reward.

\subsection{Training algorithms in a RL context}
Modern training algorithms for continuous control stochastic policies can be divided in policy gradient-based approaches and non-gradient-based approaches. The former family encompasses first order methods such as REINFORCE \cite{schulman2015high} which we will denote Vanilla Policy Gradient (VPG), approximated second order methods based on the use of natural gradient descent \cite{kakade2001natural}, local line search methods such as Trust Region Policy Optimization (TRPO) \cite{schulman2015trust}, LBGFs inspired methods such as Penalized Policy Optimization (PPO) \cite{duan2016benchmarking} and gradient free approaches such as the cross-entropy method \cite{zzzszita2006learning}.
The performance of these algorithms have been thoroughly compared in \cite{duan2016benchmarking} where the natural gradient based method Truncated Natural Policy Gradient (TNPG) and TRPO generally outperformed other approaches. In our numerical experiments we find that when the statistical patterns at the stochastic boundary conditions are stationary enough, all approaches perform conveniently. However, TNPG and TRPO outperform other methods when regime changes occur.

\subsubsection{REINFORCE}
The Reinforce algorithm \cite{schulman2015high} has been used to train our policy to maximize $\mathbb{E}\left(\eta(\pi)\right)$.

We consider a parametric policy and we denote by $\theta$ its parameters. In our case, the policy is a neural network parametrized by its weights. The input layer is filled with the environment observation, and the output layer contains the action probability distribution.

For $a \in A$, and $s \in S$, we note $\pi_\theta(a | s)$ the probability to take action $a$ while being under state $s$, and following the policy parametrized by $\theta$.

In practice, we use the following equality to compute a gradient average across multiple trajectories:
$$
\partial_\theta \mathbb{E}\left(  \eta(\pi_\theta)\right) = \mathbb{E}\left(  
\left( \sum_{t=0}^{T_{max}} \partial_\theta \log\left(\pi_\theta\left(a_t | s_t\right)\right)\right) \sum_{t=0}^{T_{max}} \gamma^t r_t
\right)
$$
The right side term can be approximated by simply running enough simulations with the given policy according to the law of large numbers.

Once we obtain the gradient $\partial_\theta \mathbb{E}\left(  \eta(\pi_\theta)\right)$ we can perform a gradient ascent on the parameters to incrementally improve our policy.

\subsubsection{Architecture and choices}
Beside these theoretical considerations and algorithms, the network architecture choice has a crucial impact on the training of the policy. If the layers are not adapted to the input data, the training algorithm will converge to a bad local minimum, or may not converge at all.

\begin{figure}[!t]
\centering
\includegraphics[trim={0 0 0 0},clip,width=3.5in]{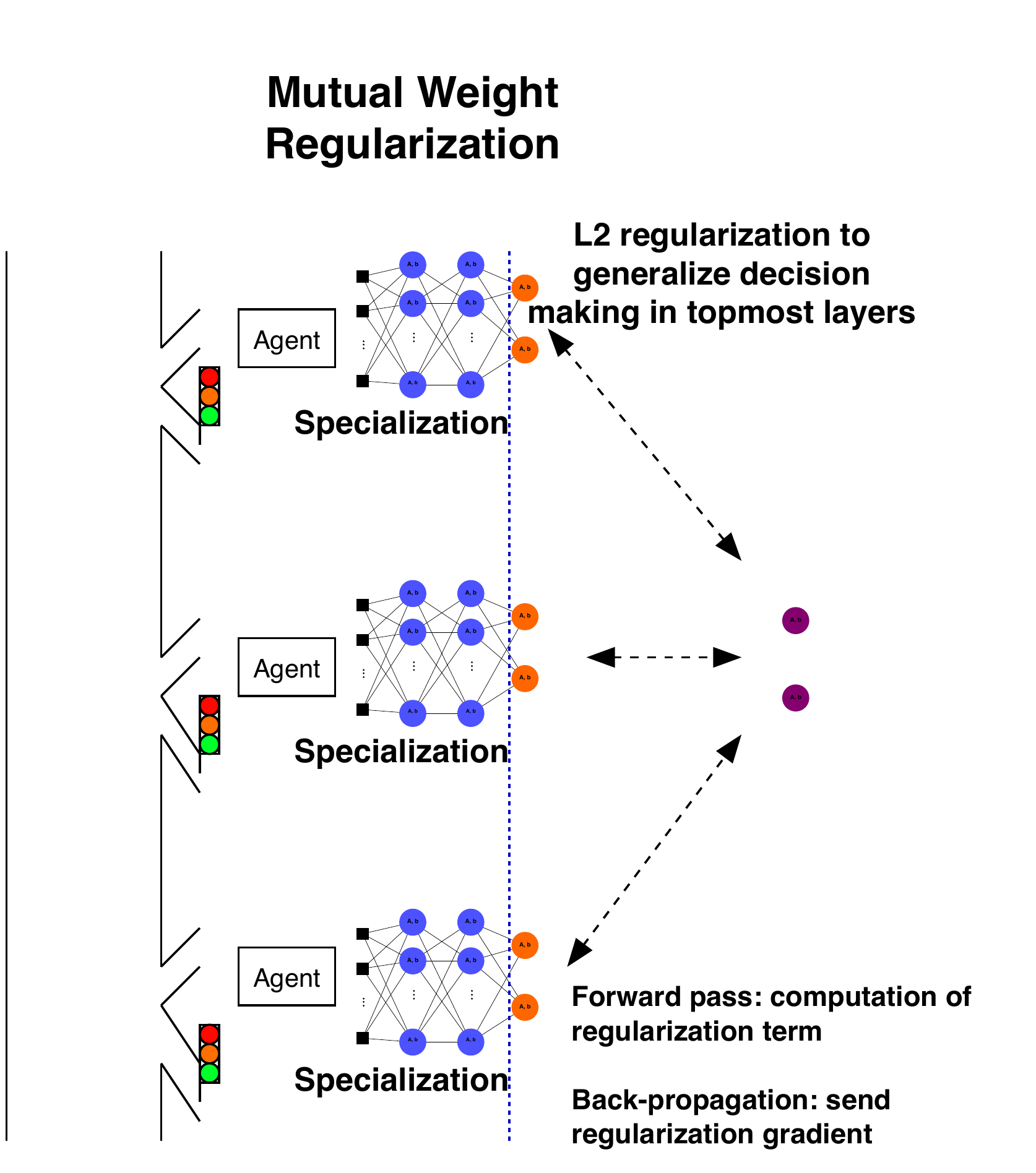}
\caption{Sharing experience across agents while allowing for specialization.}
\label{fig_share}
\end{figure}

In our settings, the observation consists of a $n \times 3$ array, where $n$ is the number of discretization cells of the highway in the simulator, and every cell contains 3 information:
\begin{itemize}
    \item The vehicle density scaled to have a median value of 0, and a standard deviation of 1 in average
    \item A boolean value indicating the presence of an off-ramp
    \item A logarithmic scaled value indicating the number of vehicle waiting in the on-ramp queue if there is any, 0 otherwise.
\end{itemize}
As this data is spatially structured, we chose to process it with convolutionnal neural layers the core idea being to handle data in a spatially invariant way. Local features are created as a function of these local values independently of the highway location. A pipeline of $3$ convolutionnal neural network layers are stacked to create local features. A last layer on the top of these convolution takes the action which consists in deciding how many vehicle can enter at the highway respective on-ramp. This practically achieve by ramp metering traffic lights. This algorithm is illustrated in Figure \ref{fig_share}.

\subsubsection{Sharing information while allowing specialization among agents: the Mutual Weight Regularization algorithm}
The features used for decision making on the low level layers of the network are created with a convolutionnal neural network, which exploits the spatial invariance of the problem.

There are two possible situations for the last layer:
\begin{itemize}
    \item Parameters sharing for all on-ramp policies. This results into having the exact same policy for all agents. This should not be the case because of local specificity of the highway, such as a reduced number of lanes, or a different speed limit for instance.
    \item Every on-ramp has its dedicated set of parameters. It allows more flexibility and different control for every on-ramp, but dramatically increases the number of parameters, does not share learning between agents, and finally does not converge to a good policy
\end{itemize}

The novel approach we introduce and call Mutual Weight Regularization (MWR) is between these 2 extremes. It acknowledges the fact that experiences and feedback should be shared between agents, to mitigate the combinatorial explosion when the number of agents scales, and still allows some agent specific modifications to adapt to local variations.

Let us consider:
\begin{itemize}
    \item $m \in \mathbb{N}$ the number of features computed per cell.
    \item $i \in R$ where $R$ is the set of cells linked to a controllable on-ramp.
    \item $X \in \mathcal{M}_{n,m}(\mathbb R)$ the output of the convolutionnal layers and we note $X[i]$ the features of the agent $i$.
\end{itemize}

We introduce $W_0, W_1, \ldots, W_n \in \mathbb{R}^m$ the distinct parameters of every agent, and define, for $i \in R$, $y_i$ as
$$y_i = \langle W_i \,, X[i]\rangle$$

The decision for the average of the distribution of actions of a given agent will be determined by a non-linear transforms of $y_i$. Similarly for the variance of this Gaussian stochastic policy distribution \cite{reinforce}.

The MWR methods consist in adding a regularization term to the global gradient used for the gradient ascent:
\begin{align}
    \partial_\theta \mathbb{E}\left( 
\eta(\pi_\theta)
- \frac{\alpha}{2} \sum\limits_{j=1}^n \left( W_j - W_{0}\right)^2
\right) = \\
\partial_\theta \mathbb{E}\left( 
\eta(\pi_\theta)
\right)
- \alpha \sum\limits_{j=1}^n \left( W_j - W_{0}\right) \partial_\theta \mathbb{E} \left( W_j - W_{0}\right)
\end{align}
Where the hyperparameter $\alpha$ defines the strength of the regularization and therefore how much mutual information is shared between agents in the gradient descent. Note that:
\begin{itemize}
    \item $\alpha = 0$ is equivalent to having independant policies for every on-ramp.
    \item $\alpha = \infty$ is equivalent to having a shared policy making algorithm for every on-ramp (shares weights).
    \item $W_0$ is not actually used for control computation, but is rather a reference weight.
\end{itemize}

\section{Experimental results}

\subsection{Proof of concept on different scenario cases}
In this first experimentation set, we demonstrate that RL can be used to control PDEs in a robust and generic way. The same training procedure converges to a successful policy for two very different tasks.
In this section, simulations are run using a simple Godunov discretization scheme for the LWR PDE (\eqnref{eq:conservationeq}), see appendix.

\subsubsection{Highway outflow control}
\textbf{Traffic management scenario.}
We consider a 5 mile section of I80W starting from the metering toll plaza and ending within San Francisco (see Figure~\ref{fig:tracking_results}. The flow is metered at the toll plaza at a rate shown in Figure~\ref{fig:tracking_results} $(iii)$ (i.e. a vehicle rate). The Godunov scheme is implemented with $20$ cells and simulated over a time span enabling several bridge crossings.
The inflow integrates random arrival rates for inbound traffic, which consists of a sinusoid perturbed by random noise. The state is vehicular density (i.e.  $\rho(t,x)$ as defined by the LWR PDE, vehicles per unit length) from which flow $F(\rho(t,x))$ (numbers of vehicles per unit time at $x$) can be computed.

\textbf{Action space and environment}
We consider the following operational scenario. The number of vehicles upstream from the meter (see Figure \ref{fig:tracking_results} $(iii)$) is randomized for each simulation in order to reproduce the diversity of freeway flow dynamics as they actually occur. 
At every time step, the observation forward propagated in the policy Neural Network is the value of current time step in time unit. An action undertaken is also a positive scalar representing the number of cars permitted to enter the highway per time unit. We train the policy to reproduce a pattern $z(t)$ on the outflow with the following reward
$$R \left( s_t, a_t \right) = -\left(\rho(t, L) - z(t)\right)^2.$$
The choice of this reward function encodes our intention to replicate the function $z(t)$ on the downstream boundary condition (located at spatial offset $L$). In particular, what is remarkable here is that this is the only way the environment provides information to the policy about the state of the freeway. Indeed, we only provide the current time step $t$ as state observation to our policy along with the reward associated with the result obtained after a simulation roll out.

\textbf{Learning in the presence of disruptive perturbations}
The discretized model we use is well-suited for representing traffic accidents. The state update mechanism may be randomly altered by accidents which drastically change the maximum flow the freeway can carry at local points in space. Accidents can be simulated by locally decreasing the maximum flow speed in a given discretization cell for a given time interval. A key goal of our work is alleviating the impact of accidents while simultaneously tracking operational objectives (e.g. desired outflow of the bridge into the city), and achieving it with a robust and generic method is a tremendous breakthrough for urban planning.
See the appendix for a presentation of the accident scenario. 

\begin{figure}
    \center
    \includegraphics[width=8cm]{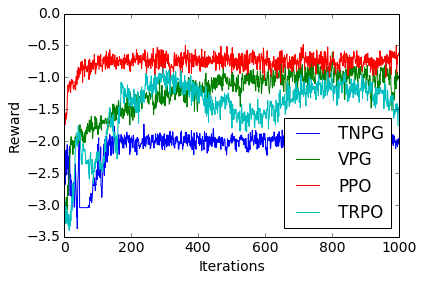}
    \caption{Comparison of different policy training algorithms for the control of the downstream boundary condition by the upstream boundary condition in the presence of accidents. We notice that TNPG performs poorly, VPG and TRPO offer mediocre performance and PPO outperforms all the methods.}
    \label{fig:rl_cmp_algos}
\end{figure}

\textbf{Learning algorithms and convergence to an effective policy}
In Figure \ref{fig:tracking_results}, we analyze the results of the control scheme learnt by a given policy consisting of two fully connected hidden layers of size $32$. The policy controls the inflow of cars (boundary control). We choose an arrival rate sufficiently high on average to provide the controller with sufficient numbers of vehicles to match the prescribed downstream conditions (note obviously in congestion this is always the case). The results prove that in spite of the problem being non-differentiable, non-smooth, non-linear and perturbed drastically by unpredictable accidents and random input queues, the policy converges to a control scheme that manages to replicate the objective density. The learning phase uses different policy update methods such as \cite{kakade2001natural,levine2013guided,schulman2015trust} and benchmarked in \cite{duan2016benchmarking}. In this benchmark, among gradient based methods, TNPG, TRPO and PPO seem to outperform the simpler REINFORCE method which only leverages first order gradient information.
In Figure \ref{fig:rl_cmp_algos} we show how PPO, TRPO and REINFORCE are all reliable in this instance and converge to more effective policies than TNPG. It is also noteworthy that PPO converges faster to a plateau of rewards.

\begin{figure}
    \centering
    \includegraphics[width=\linewidth]{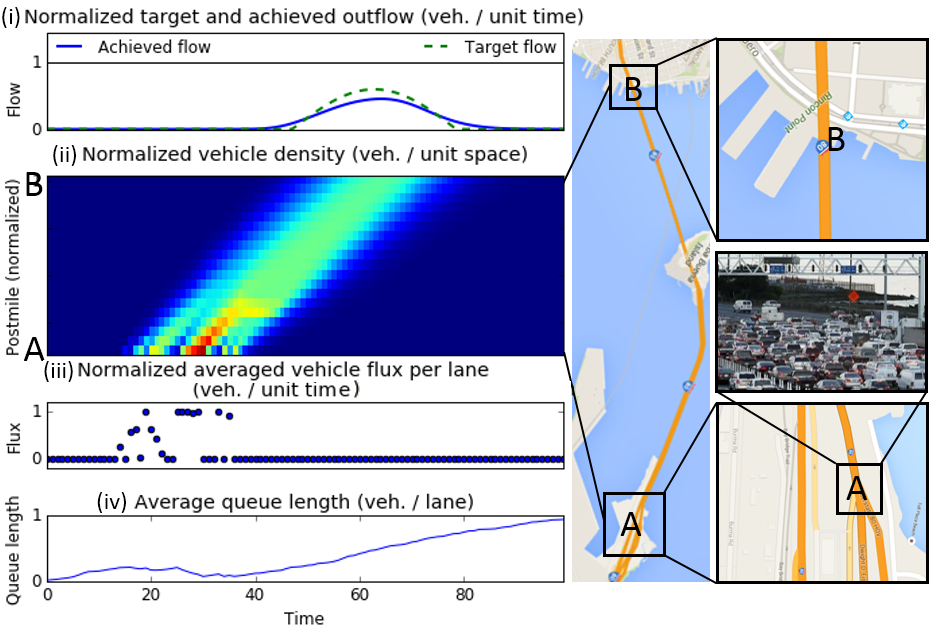}
    \caption{I80 Bay Bridge metering control. The control $(iii)$ generated by neural RL leads to a solution $(ii)$ of the PDE, which achieves an outflow shown in $(i)$, close to the desired profile in $(i)$. In the process, the queue, shown in $(iv)$ grows as vehicles into the toll plaza continue to arrive at a higher rate than allowed by the control scheme. The time space diagram i.e. $(ii)$ is a common tool used in transportation engineering to show the solution of the PDE in space ($x$) and time ($y$). The tracking policy learned through neural RL. 
    \label{fig:tracking_results}
    }
    \includegraphics[width=\linewidth]{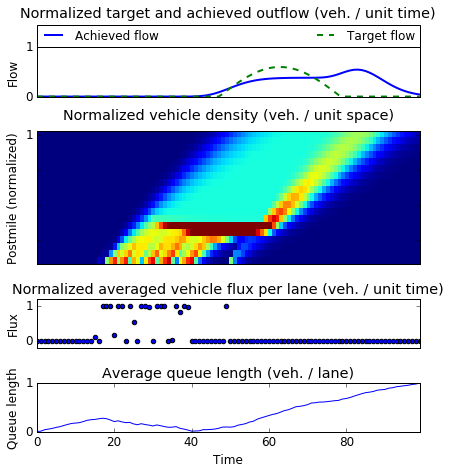}
    \caption{During the training period, major variations in the traffic model are generated to illustrate the robustness of the trained policy. Despite a highly perturbed training phase, the learnt control policy manages to converge to the prescribed objective of outbound downstream flow.
    \label{fig:tracking_results}
    }
\end{figure}

\subsubsection{Inner domain control}
\begin{figure}
  \centering
  \begin{tabular}{c}
  \includegraphics[width=0.45\textwidth]{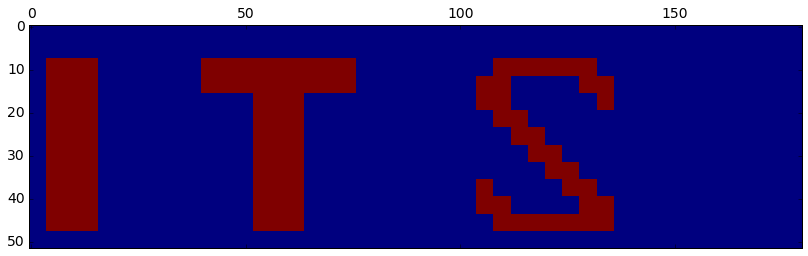}
  \\
  Objective assigned
  \\
  \includegraphics[width=0.45\textwidth]{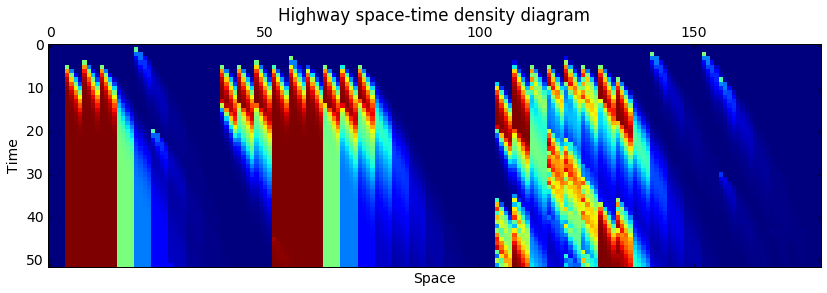}
    \\
    Objective achieved after 2000 iterations
  \end{tabular}
  \caption{Control of multiple entry points on the freeway enables the training of a policy capable of replicated a prescribed density.}
    \label{fig:control_extended}
\end{figure}

Reward shaping, in the form of assigning a target density and penalizing the $L_2$ distance between the observed density and the objective enables us to reproduce an arbitrary image with the solution density in the solution domain only by controlling it on the boundaries. The results in Figure~\ref{fig:control_extended} demonstrate the ability of the method we present to train a policy to extensively control the values of a solution to the discretized PDE we study in its solution domain. The action space here is much higher dimensional as ramps are present all along the freeway that can let cars in at a sequence spatial offsets each separated by $3$ cells. 
An off ramp split model handles the vehicle leaving the freeways. In spite of the increased dimensionality, a neural net with tree fully connected hidden layers of size $64$ trained by the TRPO method converges to a policy capable of reproducing a target solution in the interior domain as shown in Figure \ref{fig:control_extended}, whereas TNPG failed in this instance to converge to an efficient policy. From a practitioner's perspective, this example is very powerful, it shows the ability to generate arbitrary congestion patterns based on metering along the freeway. From a PDE control standpoint, this is even more powerful, as direct state actuation is a very hard problem in manufacturing and has a lot of applications with PDEs such as the (nonlinear) heat equation, Hamilton-Jacobi equations, and several others. 

\subsection{Optimal ramp metering control}
\begin{figure}[h]
    \includegraphics[trim=0.2in 3.0in 0 3.0in, width=0.5\textwidth]{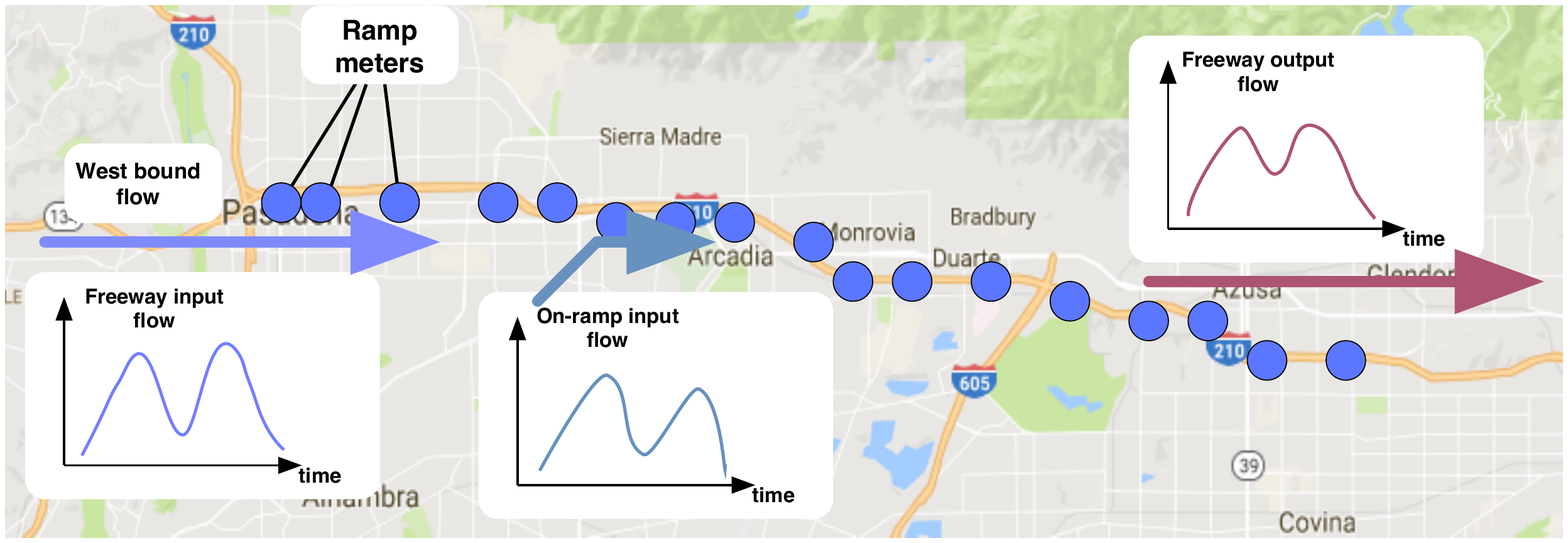}
    \caption{Setting simulated by BeATS in our experiment}
    \label{fig:freeway}
\end{figure}
\begin{figure}
    \centering
    \includegraphics[width=\linewidth]{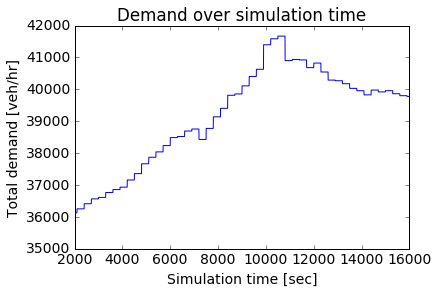}
    \caption{Total demand curve. Congestion occurs after a few hours of simulation.
    \label{fig:rm_demand}
    }
\end{figure}

In order to demonstrate the applicability of this novel method to real world cases, we consider the Ramp-Metering problem in a 20 miles (33 km) long section of the 210 Eastbound freeway in southern California as illustrated in Figure \ref{fig:freeway}.
For this simulation, we use the BeATS simulator calibrated by traffic experts based on real-world data. Every simulation run lasts for 4 hours, after a 30 minutes warmup period to initialize the freeway. The simulation starts at 12pm and the traffic peak happens between 3pm and 4pm, as the demand curve reaches a maximum (Fig. \ref{fig:rm_demand}).

Two reinforcement learning algorithms and the ALINEA control scheme are benchmarked against the baseline scenario in which no control occurs at all:
\begin{itemize}
    \item \textbf{NoRM, baseline}: The baseline without any ramp metering. Cars instantly enter the freeway when they reach an onramp, and if the freeway has enough capacity
    \item \textbf{NoMWR, standard deep reinforcement learning}: The Reinforcement Learning based policy we introduce, trained with shared weights for the last layer
    \item \textbf{MWR, novel approach to training}: The same policy as NoMWR, but trained with MWR.
    \item \textbf{Alinea, parametric control}: The state of the art reference algorithm, using with model and parameters the exact same values used in the simulator it is being benchmarked on.
\end{itemize}

\subsubsection{Reinforcement Learning problem}
In this scenario, the agent takes an action every 32 seconds. An action is a $\mathbb{R}^{29}$ vector with the ramp-metering rates for the 29 on-ramps on the highway section, in vehicle per time unit.
The reward we collect at every time step is the total outflow in the last 32 seconds (in number of vehicle per time unit).

The highway is discretized in 167 cells of 200 meters each to generate an observation vector in $\mathbb{R}^{167 \times 3}$. For every cell, the following 3 data values are provided to the network:
\begin{itemize}
    \item Density in vehicle per space unit, normalized
    \item A boolean value indicating the presence of an off-ramp on this cell
    \item The number of cars waiting in this cell's onramp queue (logarithmically scaled), or 0 if there is no onramp
\end{itemize}

It is worth mentioning that the Reinforcement Learning policies are trained in a stochastic way to ensure that a generic policy is learned. This is done by introducing some noise in the actions taken by the Neural Network: the ramp metering actually applied is sampled from a Gaussian distribution centered on the network output. This strategy, along with the use of shared learning techniques over the 29 onramps, globally prevents overfitting issues.

\subsubsection{Numerical results}
\begin{figure}
    \centering
    \includegraphics[width=\linewidth]{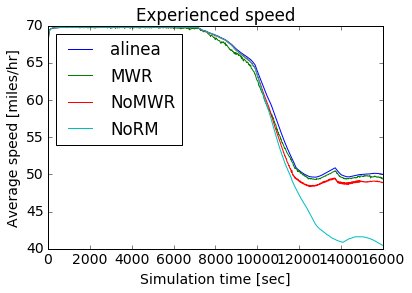}
    \caption{Alinea maintains speed above the optimal level defined during calibration by experts. Reinforcement Learning methods and the MWR especially, manage to implicitly deduce this optimal speed during training.
    \label{fig:rm_exp_speed}
    }
\end{figure}

\begin{figure}
    \centering
    \includegraphics[width=\linewidth]{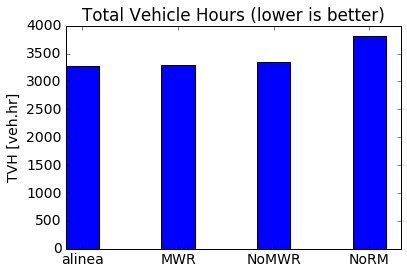}
    \caption{Total Vehicle Hour during congestion time
    \label{fig:rm_tvh}
    }
\end{figure}

\begin{figure}
    \centering
    \includegraphics[width=\linewidth]{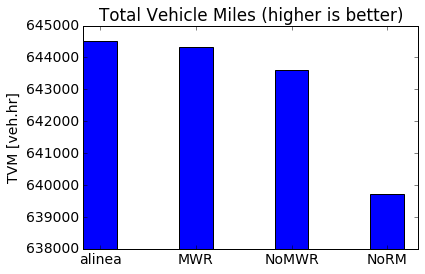}
    \caption{MWR provides a significant performance improvement and almost reaches Alinea performance. Compared to no ramp-metering, MWR reproduces 96\% of Alinea performance improvement, while the regular RL training only reaches 80\% of Alinea performance.
    \label{fig:rm_tvm}
    }
\end{figure}

After the training, we compared the results of our approach to existing algorithm on several criteria. 
The average speed is globally increased as expected (Fig. \ref{fig:rm_exp_speed}). We also report the Total Vehicle Miles in Fig. \ref{fig:rm_tvm}, along with the Total Vehicle Hour (Fig. \ref{fig:rm_tvh}) that assess the performance of our approach with a single score. In both cases, the MWR training approach provides a significative performance increase regular parameters sharing (NoMWR), and almost reaches the performance of the reference and state of the art parametric method Alinea.

\begin{table}[]
    \centering
    \begin{tabular}{c c c}
        Approach & Score in veh.hr & Score in veh.mile  \\
         &  (lower is better)  & (higher is better) \\
        ALINEA & 10514 & 644522 \\
        MWR & 10575 & 644334 \\
        NoMWR & 10617 & 643605 \\
        NoRM & 11085 & 639709
    \end{tabular}
    \caption{Aggregated scores of the different control strategies over the congested period. Reinforcement learning enables a non parametric control scheme whose performance is similar to that of the parametric ALINEA scheme and MWR, the new learning algorithm we introduce, for improves this performance.}
    \label{tab:agg_results}
\end{table}

\begin{figure}
    \centering
    \includegraphics[width=\linewidth]{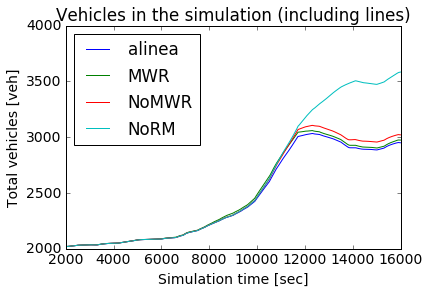}
    \caption{RM methods reduce by 20\% the number of vehicles in the freeway during congestion peak times. Our RL learning trained with MWR maintains the number of vehicles in the simulation to a level similar to what Alinea performs.
    \label{fig:rm_tot_veh}
    }
\end{figure}
\begin{figure}
    \centering
    \includegraphics[width=\linewidth]{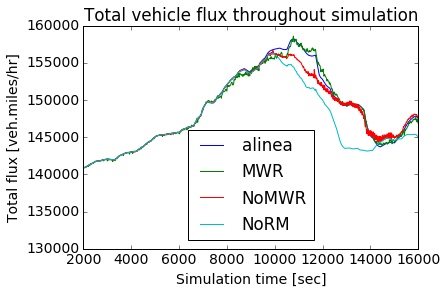}
    \caption{Vehicle flux
    \label{fig:rm_veh_flux}
    }
\end{figure}

\section{Conclusion}
We have demonstrated how neural RL 
substantially improves upon the state-of-the-art in the field of control of discretized PDE. It enables reliable non-parametric control while offering theoretical guarantees similar to that of classic parametric control techniques.
In particular, neural RL can be applied without an explicit model of system dynamics, and instead only requires the ability to simulate the system under consideration. 
Through our experimental evaluation, we demonstrated that neural RL approach can be used to control discretized macroscopic vehicular traffic equations by their boundary conditions in spite of accidents drastically perturbing the system. 
Achieving such robustness is a significant breakthrough in the field of control of cyberphysical systems. Specific to the practice of transportation, the results are a major disruption as they enable us to beat current controllers by performing adaptive control, without the need for model calibration. 
By eliminating the need for calibration, our method addresses one of the critical challenges and dominant causes of controller failure making our approach particularly promising in the field of traffic management.
We also introduced a novel algorithm, MWR, to achieve multi-agent control and leverage trial and error experiences across different agents while at the same time allowing each agent to learn how to tailor its behavior to its localization in the large cyberphysical system under study.


%

\appendices
\section{Godunov discretization scheme}
\label{sec:godunovDicr}
Because of the presence of discontinuities in their solutions, the benchmark PDE we consider is formulated in the weak sense.
We consider an open set $\Omega$ in $\mathbb{R}$, and a measurable function $\rho=\rho(t,x)$ is a distributional solution to the system of conservation laws
\begin{equation}
    \partial_t \rho + \partial_x F \left( \rho \right) = 0.
    \label{eq:conservationeq}
\end{equation}
if, for every $C^1$ function $\phi$ defined over $\Omega$ with compact support, one has
\begin{equation}
    \int_{t \in \mathbb{R}, x \in \Omega} \left[\rho \partial_t \phi + F \left( \rho \right) \partial_x \phi \right] dx dt = 0.
    \label{eq:weaksolution}
\end{equation}
The operator $F$ in \eqnref{eq:conservationeq} will be referred to as flux function, also called the ``Fundamental diagram'' in transportation engineering. The operator $F$ defines entirely the dynamics at stake and therefore is often domain specific. Given an initial condition 
\begin{equation}
    \rho(0,x) = \bar{\rho}(x)
    \label{eq:weakinit}
\end{equation}
where $\bar{\rho}(x)$ is locally integrable,
$y : \left[0, \: T\right] \times \mathbb{R} \rightarrow \mathbb{R}$ is a distributional solution to the Cauchy problem defined by \eqnref{eq:weaksolution} and \eqnref{eq:weakinit} if 
\begin{equation}
    \int_{t \in [0, \: T], x \in \Omega} \left[\rho \partial_t \phi + F \left( \rho \right) \partial_x \phi \right] dx dt 
    +
    \int_{x \in \Omega} \left[\bar{\rho}(x) \phi(0, x) \right] dx
    = 0.
    \label{eq:weakCauchy}
\end{equation}
for every $C^1$ function $\phi$ with compact support contained in $\left]-\infty,\:T\right] \times \mathbb{R}$. If $y$ is a continuous function from $\left[0, \: T \right]$ into the set of locally integrable on $\Omega$, solves the Cauchy problem above in the distribution sense, $\rho$ is referred to as a weak solution to the Cauchy problem. 

The Dirichlet problem corresponding to a boundary condition (as done later in the article) can be formulated in a similar manner and is left out of the article for brevity. Such a definition of weak solutions is not sufficient to guarantee their being admissible solutions.
The entropy condition
guarantees the uniqueness to the problem and continuous dependence with respect to the initial data (derivation also left out of the article for brevity). 

The Godunov's scheme computes an approximate weak solution $\tilde{\rho}$ to the Dirichlet problem \eqnref{eq:weaksolution}, \eqnref{eq:weakinit} with the following recursive equation and Godunov flux $G$:

\footnotesize{
\begin{align}
    \tilde{\rho}(n \Delta t + \Delta t, i \Delta x) = \nonumber\\ \tilde{\rho}(n \Delta t, i \Delta x)
    - \frac{\Delta t}{\Delta x}
    G\left(
    \tilde{\rho}(n \Delta t, i \Delta x - \Delta_x),
    \tilde{\rho}(n \Delta t, i \Delta x + \Delta_x)
    \right) \nonumber
\end{align}
\begin{align}
    G \left(\tilde{\rho}_l, \tilde{\rho}_r\right) = & F(\tilde{\rho}_l) \text{ if } \tilde{\rho}_l > \tilde{\rho}_r \text{ and } ({F(\tilde{\rho}_r) - F(\tilde{\rho}_l)})/({\tilde{\rho}_r - \tilde{\rho}_l}) > 0, \nonumber \\
    & F(\tilde{\rho}_r) \text{ if } \tilde{\rho}_l > \tilde{\rho}_r \text{ and } ({F(\tilde{\rho}_r) - F(\tilde{\rho}_l)})/({\tilde{\rho}_r - \tilde{\rho}_l}) < 0, \nonumber \\
    & F(\tilde{\rho}_l) \text{ if } \tilde{\rho}_l < \tilde{\rho}_r \text{ and } DF(\tilde{\rho}_l) > 0, 
    \nonumber \\
    & F(\tilde{\rho}_l) \text{ if } \tilde{\rho}_l < \tilde{\rho}_r \text{ and } DF(\tilde{\rho}_r) < 0,
    DF^{-1}(0) \text{ otherwise }.
    \nonumber
    \label{eq:explicit}
\end{align}
}

The Godunov scheme is second order accurate in space. Unfortunately, like most numerical schemes, it is non-differentiable because of the presence of the ``if-then-else'' statements in its explicit form.
Another problematic aspect related with computing numerical weak entropy solutions with most numerical schemes (incl. Godunov) is their relying on a numerical evaluation of $F$ which often takes the form of a parametrized function. The estimation of these parameters is often difficult and it is practically intractable to assess the impact of the parameter uncertainty on the solutions because of the non-linearity, non-smoothness and non-differentiability of the schemes.




\ifCLASSOPTIONcaptionsoff
  \newpage
\fi



%
\bibliographystyle{IEEEtran}
\bibliography{biblio}

%

\begin{IEEEbiographynophoto}{Francois Belletti}
Francois Belletti graduated in 2013 from Ecole Polytechnique - France, with a MSc in Probability and Finance. He later graduated with a MSc in Distributed Computing from Imperial College London before starting a PhD in Computer Science at UC Berkeley. His work focuses on scalable computing for the analysis and control of time series.
\end{IEEEbiographynophoto}

\begin{IEEEbiographynophoto}{Daniel Haziza}
 received the B.E. and M.E. degrees in Computer Science with a minor in Applied Mathematics from the Ecole Polytechnique, France, in 2014 and 2016 respectively.
In 2016, he was doing research in the Bayen Lab at UC Berkeley. His research interests include Machine Learning, Data Science and its applications.
\end{IEEEbiographynophoto}

\begin{IEEEbiographynophoto}{Gabriel Gomes} is an Assistant Research Engineer with the Institute for Transportation Studies at the University of California at Berkeley. He holds a doctoral degree in systems and control theory from the Department of Mechanical Engineering at UC Berkeley. Dr. Gomes' work focuses on modeling, control, and simulation of transportation systems.
\end{IEEEbiographynophoto}


\begin{IEEEbiographynophoto}{Alexandre M. Bayen}
Alexandre Bayen received the Engineering Degree in applied mathematics from the Ecole Polytechnique, France, in July 1998, the M.S. degree in aeronautics and astronautics from Stanford University in June 1999, and the Ph.D. in aeronautics and astronautics from Stanford University in December 2003. He was a Visiting Researcher at NASA Ames Research Center from 2000 to 2003. Between January 2004 and December 2004, he worked as the Research Director of the Autonomous Navigation Laboratory at the Laboratoire de Recherches Balistiques et Aerodynamiques, (Ministere de la Defense, Vernon, France), where he holds the rank of Major. He is currently an Associate Chancellor Professor, and has been the Director of the Institute for Transportation Studies (ITS) since 2014.
\end{IEEEbiographynophoto}




\end{document}